\documentclass[sigconf]{acmart}
\usepackage{balance}
\AtBeginDocument{%
  }

\copyrightyear{2025}
\acmYear{2025}
\setcopyright{acmlicensed}\acmConference[McGE '25]{Proceedings of the 3rd International Workshop on Multimedia Content Generation and Evaluation: New Methods and Practice}{October 27--28, 2025}{Dublin, Ireland}
\acmBooktitle{Proceedings of the 3rd International Workshop on Multimedia Content Generation and Evaluation: New Methods and Practice (McGE '25), October 27--28, 2025, Dublin, Ireland}
\acmDOI{10.1145/3746278.3759378}
\acmISBN{979-8-4007-2060-4/2025/10}
\begin{document}

\title{Moyun: A Diffusion-Based Model for Style-Specific Chinese Calligraphy Generation}



\author{Kaiyuan Liu}
\affiliation{
  \institution{School of Computer Science and Technology, \\ East China Normal University}
  \city{Shanghai}
  \country{China}
}
\email{liukaiyuan@stu.ecnu.edu.cn}
\orcid{0000-0002-2640-8026}

\author{Jiahao Mei}
\affiliation{
  \institution{School of Computer Science and Technology, \\ East China Normal University}
  \city{Shanghai}
  \country{China}
}
\email{jhmei@stu.ecnu.edu.cn}

\author{Hengyu Zhang}
\affiliation{
  \institution{School of Computer Science and Technology, \\ East China Normal University}
  \city{Shanghai}
  \country{China}
}
\email{hyzhang@stu.ecnu.edu.cn}

\author{Yihuai Zhang}
\affiliation{
  \institution{School of Computer Science and Technology, \\ East China Normal University}
  \city{Shanghai}
  \country{China}
}
\email{yhzhang@stu.ecnu.edu.cn}

\author{Daoguo Dong}
\affiliation{
  \institution{School of Computer Science and Technology, \\ East China Normal University}
  \city{Shanghai}
  \country{China}
}
\email{dgdong@cs.ecnu.edu.cn}
\authornote{Corresponding author. E-mail:dgdong@cs.ecnu.edu.cn}

\author{Liang He}
\affiliation{
  \institution{School of Computer Science and Technology, \\ East China Normal University}
  \city{Shanghai}
  \country{China}
}
\email{lhe@cs.ecnu.edu.cn}


\begin{abstract}
    Although Chinese calligraphy generation has made progress in style transfer or font synthesis,
    generating calligraphy by precisely specifying the calligrapher, font, and character style remains challenging.
    To address this, we propose a new Chinese calligraphy generation model ``Moyun'' ,
    which replaces the UNet in the Diffusion model with Vision Mamba and introduces the 
    TripleLabel control mechanism to achieve controllable calligraphy generation. 
    The model was tested on our large-scale dataset ``Mobao'', which contains over 1.9 million images. 
    The results demonstrate that ``Moyun'' can effectively control the generation process and produce calligraphy in the specified style. 
    Even for calligraphy the calligrapher has not written, ``Moyun'' can generate calligraphy that matches the style of the calligrapher.
\end{abstract}

\ccsdesc[300]{Computing methodologies~Artificial intelligence}
\keywords{Chinese Calligraphy, Diffusion Model, Mamba}


\maketitle

\section{Introduction}
\label{sec:intro}

\begin{figure}[t]
    \centering
    \includegraphics[width=1 \linewidth]{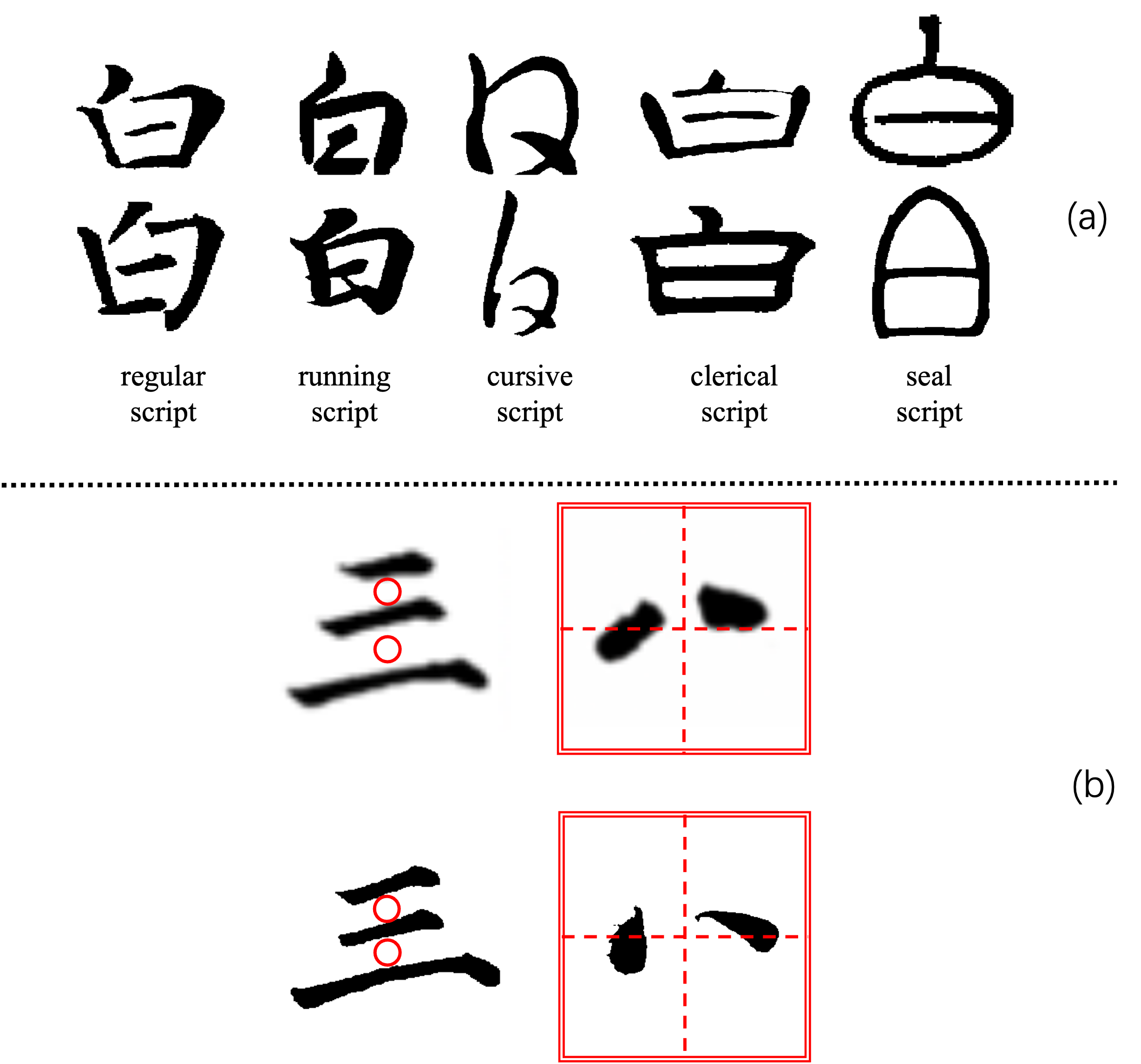}
    \caption{(a) shows the character ``bai'' (which means ``white'' in Chinese) written in different fonts by various calligraphers. Each column represents a different font, and each row corresponds to a different calligrapher. (b) The first row shows calligraphy generated by Calliffusion, while the second row shows the ground truth. In the first column (san, regular script, Yan Zhenqing), the strokes in the ground truth are evenly spaced, but Calliffusion's result is not. In the second column (ba, cursive script, Su Shi), Calliffusion's output appears less stable than the ground truth from an aesthetic standpoint.}
    \label{fig:intro}
\end{figure}

Chinese calligraphy, with a history spanning over thousands of years, is a cherished cultural treasure of China which represents the artistic of Chinese characters handwriting. As a world intangible cultural heritage, its unique aesthetic characteristics are reflected in the lightness and heaviness of the brushstrokes, the thickness and dryness of the ink levels, and the virtual and real echoes of the layout.
Chinese calligraphy is rich in variations. Chinese has tens of thousands of characters, each with a different meaning. Additionally, a single character can be written in various fonts, such as regular script, running script, cursive script, clerical script, seal script, and so on. 
Moreover, the writing of the same font differs between calligraphers, as shown in Figure \ref{fig:intro}a. The writing of the same font by the same calligrapher shows consistency, which we refer to as calligraphic style.
The rich variations in Chinese calligraphy require an extended period of study for an average person to master. However, Chinese calligraphy is widely used, which has led people to explore the use of AI for generating Chinese calligraphy.
Recently, GAN and Diffusion models are applied to Chinese calligraphy generation \cite{calligan,zigan,calliffusion,zi2zi}, yielding impressive outcomes.

However, there are still some problems with the current generation models of Chinese calligraphy.

ZiGAN \cite{zigan} is a seminal work that applies GAN to the field of Chinese calligraphy generation. ZiGAN constructed several small-scale datasets, each containing calligraphy in a consistent style, and trained models separately on them. As a result, the trained models gained the ability to transform standard printed characters into calligraphy\footnote{In the following text, we use ``Calligraphy'' to refer to a single character image generated by the calligraphy generation model.} with the style of the specific dataset used during training.  However, it cannot generate calligraphy guided by specific calligraphers, fonts, and characters. 
Calliffusion \cite{calliffusion} is the first Chinese calligraphy generation model based on diffusion model\cite{DDPM}. Calliffusion uses descriptive text involving ``character, font, and calligrapher'' to effectively guide the generation process. However, it relies on a BERT\cite{BERT} model pre-trained on Chinese language dataset as the text encoder, which struggles to accurately understand domain-specific terms in calligraphy, such as the names of calligraphers. For example, a calligrapher's name might be split into two tokens. Additionally, this approach introduces extra computational cost, which is unnecessary for calligraphy generation tasks. To better accomplish this task, we proposed a multilabel mechanism, where independent classification labels correspond to the calligrapher, font, and character, and these labels are combined to control the generation process.
In this way, Moyun can not only reduce the computational cost, but also control the generated calligraphy more accurately.

Calliffusion uses an Diffusion architecture based on Unet\cite{unet}, but Unet does not adequately fit the structural relationships between the strokes of the characters, which is shown in Figure \ref{fig:intro}b. 
The convolution operation of UNet is limited by the size of the receptive field, making it difficult to model the complex topological relationships between calligraphy strokes (such as the flying white connection and the blank space in the cursive script).
Our model is based on the Vision Mamba\cite{visionmamba}, which processes images through patchify. Additionally, we incorporated the more efficient Mamba model\cite{mamba2}. Experiments demonstrate that our model provides a better fit to the structure between strokes.
Vision Mamba implements long-range dependency modeling of pixel sequences through the state-space model (SSM), and its hardware-aware algorithm design is particularly suitable for processing brushstroke associations across multiple scales in calligraphy images.

The ZiGAN's dataset\cite{zigan} is relatively small, containing only 9 sub-datasets, with each sub-dataset consisting of 6,000 images of a single style. Moreover, this calligraphy dataset is not currently open to the public.
In contrast, we collected 1.9 million single-character calligraphy images with diverse styles and detailed annotations. However, these calligraphy works come from different sources and contain a lot of noise. 
For example, cracks on the inscriptions, broken paper, extra ink spots, and even incomplete strokes caused by dry brush strokes.
Due to varying collection conditions and different noise distributions, traditional binarization methods performed poorly. To address this, we designed a new  binarization method based on SAM\cite{SAM}, which achieved better results.
This method uses the prompt mechanism of Segment Anything Model to guide the segmentation of calligraphy subjects through interactive annotation, solving the problem of binarization under complex backgrounds.

In summary, we have the following contributions:
\begin{itemize} 
\item We propose a Chinese calligraphy model called ``Moyun'', which is capable of generating $ 256 \times 256 $ single character calligraphy images. Moyun produces stroke structures and brushstrokes that align with those of real calligraphy, achieving state-of-the-art (SOTA) quality in image generation. 
\item We innovatively introduced a multilabel calligraphy generation control mechanism ``TripleLabel'', which allows for generating characters with label of calligrapher, font and character. Additionally, it can generate characters that the calligrapher has never written before. 
\item We constructed a large-scale, well-annotated Chinese calligraphy dataset ``Mobao'' containing more than 1.9 million high-quality binarized images using SAM. 
\end{itemize}

\section{Related Work}
\label{sec:related}


\textbf{Generative Adversarial Networks}
Generative Adversarial Networks (GANs)~\cite{GAN} represent a powerful framework in unsupervised learning, consisting of two competing neural networks: a generator and a discriminator. The generator aims to synthesize data that closely resembles real data by mapping from a random noise vector (typically drawn from a Gaussian or uniform distribution) to the data space. Conversely, the discriminator functions as a binary classifier, tasked with distinguishing between real data samples from the training set and synthetic data produced by the generator. The two networks are trained in an adversarial manner, engaging in a minimax game where the generator seeks to maximize the probability of the discriminator making an incorrect classification, while the discriminator strives to improve its accuracy. This dynamic leads to a progressive enhancement in the quality of generated samples as training proceeds.
Since its introduction in 2014, GAN has made breakthrough progress in the field of image generation. In 2015, DCGAN \cite{DCGAN} achieved stable training of deep convolutional GAN for the first time through a fully convolutional network architecture and a specific normalization strategy.
At the application level, GAN has gone beyond the scope of image generation and has spawned diverse variants such as image-to-image translation (Pix2Pix \cite{pix2pix}), unsupervised domain transfer (CycleGAN \cite{CycleGAN}), and high-resolution face synthesis (StyleGAN \cite{StyleGAN2,StyleGAN}).
Research on GAN-based calligraphy generation has been ongoing for a long time, with the earliest work being zi2zi \cite{zi2zi}, an open-source project based on pix2pix \cite{pix2pix}. 
After that, there were calliGAN \cite{calligan}, ZiGAN\cite{zigan}, and end-to-end\cite{zhou_end--end_2021} models. 
The most recent work is a style transfer model called CCST-GAN\cite{ccstgan}.\\
\textbf{Diffusion Model}
The Diffusion model has emerged as one of the most widely studied and influential frameworks in image generation due to its strong generative capabilities and theoretical elegance. The foundational structure of modern diffusion models was established by the Denoising Diffusion Probabilistic Models (DDPM)~\cite{DDPM}, which formalized the process of gradually adding Gaussian noise to data (forward diffusion) and then learning to reverse this process to generate samples (reverse diffusion). This paradigm demonstrated high sample quality and training stability compared to earlier generative models such as GANs. Subsequently, the Denoising Diffusion Implicit Models (DDIM)~\cite{DDIM} improved upon DDPM by introducing a non-Markovian reverse process, enabling significantly faster sampling without retraining, thus enhancing the practicality of diffusion models. 
Further advancements were achieved with Latent Diffusion Models (LDM)~\cite{LDM}, which introduced a Variational Autoencoder (VAE)~\cite{VAE} to learn a compressed latent space representation of images. By performing the diffusion process in this lower-dimensional latent space rather than in pixel space, LDM drastically reduced computational cost and memory usage while maintaining high-fidelity generation, making it feasible to scale diffusion models to high-resolution images and diverse domains such as text-to-image synthesis. 
A major architectural innovation came with the introduction of Diffusion Transformers (DiT)~\cite{DiT}, which replaced the traditional U-Net backbone commonly used in diffusion models with a Vision Transformer (ViT)~\cite{transformer}-based architecture. DiT demonstrated that Transformers, with their global attention mechanisms, can serve as powerful backbones for generative modeling, achieving competitive or superior performance compared to convolutional counterparts and highlighting the scalability of diffusion models with increased model capacity. 
Despite the rapid progress in general image generation, research on diffusion-based Chinese calligraphy generation remains relatively nascent. Early efforts in calligraphy synthesis primarily relied on GANs or rule-based systems, but recent works have begun to explore the potential of diffusion models to capture the delicate stroke structures, stylistic variations, and artistic nuances inherent in calligraphic art. The application of diffusion models in this domain presents unique challenges, including the need for fine-grained control over stroke dynamics and preservation of cultural authenticity, but also offers promising opportunities for digital heritage preservation and creative design. 
DP-Font\cite{dpfont} is a Chinese font generation method based on diffusion model and physical information neural network (PINN). It improves the realism and structural accuracy of the generated fonts by introducing stroke order constraints and physical equations to guide the learning process. Compared with existing methods, DP-Font shows better performance in generating high-quality and personalized Chinese fonts.
However, this work focuses on font generation rather than calligraphy generation, and the font generation task cannot focus on the style of the calligrapher.
The earliest work on calligraphy is Calliffusion\cite{calliffusion}, which pioneered the application of diffusion models to Chinese calligraphy generation by employing a UNet-based architecture.
Calliffusion is capable of generating calligraphy in five major fonts, and can effectively mimic the unique styles of famous calligraphers.
To enable controllable generation, the model incorporates a text encoder that processes a fixed-format textual prompt describing the desired character, script, and style.
These conditions are then integrated into the diffusion process through cross-attention mechanisms, allowing precise control over the generated output.
Compared with Callifusion, CalliffusionV2\cite{calliffusionV2} further introduces reference images to achieve stroke-level detail adjustment. 
The system adds multi-modal control capabilities and uses LoRA fine-tuning technology to quickly adapt to new styles with only 5 samples, more accurately restoring the natural strokes of complex calligraphy.

\section{Methodology}
\label{sec:methodology}

\begin{figure}[t]
    \centering
    \includegraphics[width=1\linewidth]{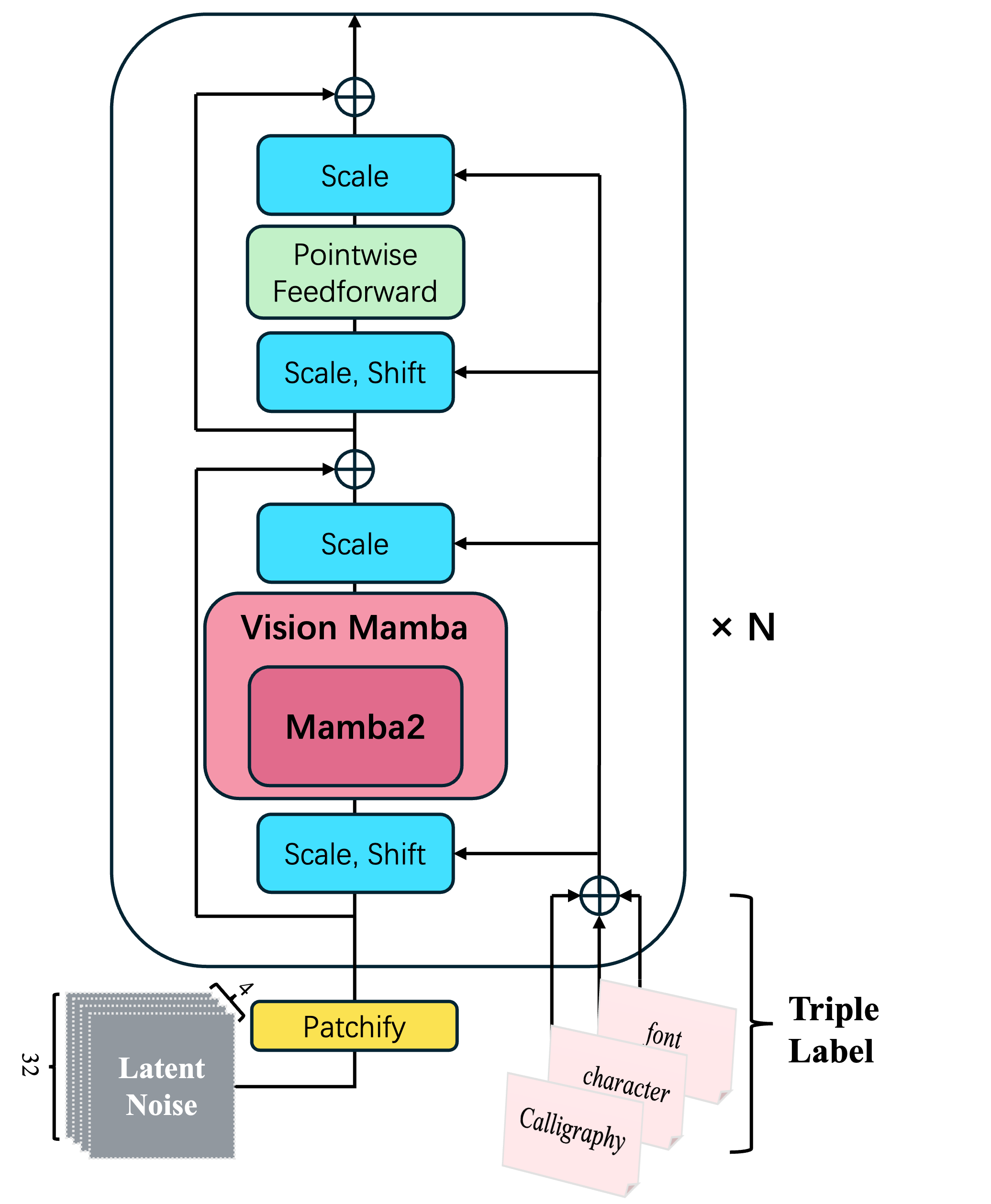}
        \caption{``Moyun'' architecture. The input latent noise is patched. The label is a combination of the calligrapher, font, and character. We used Mamba2-Replacement-Vision Mamba to process the patches.}
    \label{fig:model}
    \vspace{-15pt}
\end{figure}

Figure \ref{fig:model} shows the architecture of ``Moyun''. ``Moyun'' is a diffusion model based on Vision Mamba\cite{visionmamba}, optimized with Mamba2\cite{mamba2}. Furthermore, we control the generation process by the TripleLabel mechanism.

\subsection{Preliminaries}
Moyun is a diffusion model, so our training and inference follow the diffusion \cite{DDPM,DDIM} approach.
The forward process gradually adds noise to the real data, specifically to our calligraphy image \(x_0\). The diffusion process follows the equation as shown in \eqref{equ:1} .
\begin{equation}
\label{equ:1}
    q(x_t|x_0) = \mathcal{N}\left(x_t; \bar{\alpha}_t x_0, (1 - \bar{\alpha}_t) \mathbf{I} \right)
\end{equation}
Furthermore, the noise at step \(t\), \(x_t\) can be obtained by equation 
\begin{equation}
\label{equ:2}
    x_t = \bar{\alpha}_t x_0 + \sqrt{1 - \bar{\alpha}_t} \, \epsilon_t
\end{equation}
The diffusion model trains the reverse denoising process, and the denoising equations are as follows:
\begin{equation}
    p_{\theta}(x_{t-1} \mid x_t) = \mathcal{N}(\mu_{\theta}(x_t), \Sigma_{\theta}(x_t))
\end{equation}
Our loss function is given by the following equation:
\begin{equation}
\mathcal{L}_{\text{simple}}(\theta) = \| \epsilon_\theta (x_t) - \epsilon_t \|^2
\end{equation}
\subsection{Model and Block Design}

The proposed architecture of moyun model follows a three-stage hierarchical structure:(1) Encoder , (2) Patchify Module , and (3) Iterative Structural Refinement Blocks ($\times$N) . This design ensures explicit separation of spatial-to-sequence conversion and calligraphy feature extraction.\\
\textbf{Encoder}
The input image $\mathbf{X} \in \mathbb{R}^{a \times a \times C}$ (where $C$ is the channel dimension) is first encoded into a compact latent representation through a pre-trained variational autoencoder (VAE)\cite{VAE}.
The encoder network $\mathcal{E}$ maps the image to a latent space:
$$
\mathbf{Z} = \mathcal{E}(\mathbf{X}) \in \mathbb{R}^{b \times b \times D}
$$
where $b \times b$ is the spatial resolution of the latent space, and $D$ is the channel dimension.\\
\textbf{Patchify Module}
The latent tensor $\mathbf{Z}$ undergoes patchification to convert spatial information into a sequence of tokens:
$$
\{\mathbf{z}_i\}_{i=1}^{n} = \text{Patchify}(\mathbf{Z}) \in \mathbb{R}^{n \times d}
$$
where each patch $\mathbf{z}_i \in \mathbb{R}^{p \times p \times D}$ is flattened into a token $\mathbf{z}_i \in \mathbb{R}^{d}$. 
The number of patches $n = \frac{b^2}{p^2}$ is determined by the patch size $p$.\\
\textbf{moyun Block}
To enhance the model's capability in capturing hierarchical structural patterns inherent in calligraphic artworks, we propose a novel block design that integrates state-of-the-art sequence modeling architectures with diffusion-based generative frameworks. 
The core innovation lies in replacing the conventional U-Net backbone of diffusion models with Vision Mamba\cite{visionmamba}, while incorporating the advanced Mamba2\cite{mamba2} architecture for optimized contextual learning. 
Thus, each patch corresponds to a small part of the calligraphic structure, and Mamba's efficient contextual relational ability effectively matches the relationships between patches, thereby better the model learning the structure of the calligraphy.
In total, the entire moyun Block is iterated $N$ times.

\subsection{TripleLabel control}
To address the unique multi-conditional requirements in Chinese calligraphy generation, which include calligrapher, font, and character, we propose the TripleLabel control mechanism - 
a lightweight yet effective framework enabling simultaneous control over three critical aspects: calligrapher style , font structure , and character semantics.
This design addresses three key challenges in calligraphy generation: (1) combinatorial generalization across calligrapher-font-character combinations, 
(2) efficient condition encoding without additional encoder overhead, and (3) precise condition during diffusion.\\
Through the TripleLabel method, each calligrapher, font, and character is mapped to a unique class label, represented by a number, and each label is Independent from others. 
This allows the combination of labels to generate calligraphy that the calligrapher has not written before. 
For example, in the training set, there is no regular script ``bai'' written by calligrapher Wang Xizhi, but there is an calligraphy of ``bai'' written by another calligrapher. 
During the generation process, it is still possible to generate regular script ``bai'' characters in the style of Wang Xizhi  by providing the combinations labels of ``Wang Xizhi'', ``regular script'' and ``bai''.\\
In the model, the input labels are transformed into the corresponding embedding vector through three separate trainable embedding tables.
$$
\begin{aligned}
\mathbf{e}_{\text{calli}} &= \text{Embed}_{\text{calli}}(l_{\text{calli}}) \in \mathbb{R}^{d_e} \\
\vspace{5pt}
\mathbf{e}_{\text{font}} &= \text{Embed}_{\text{font}}(l_{\text{font}}) \in \mathbb{R}^{d_e} \\
\vspace{5pt}
\mathbf{e}_{\text{char}} &= \text{Embed}_{\text{char}}(l_{\text{char}}) \in \mathbb{R}^{d_e}
\end{aligned}
$$
where $l_{*}$ denotes class labels (e.g., $l_{\text{calli}} \in \{1,2,...,K_{\text{calli}}\}$).\\
This disentangled design leverages the additive property of Euclidean spaces for compositional control:
$$
\mathbf{e}_{\text{total}} = \mathbf{e}_{\text{calli}} + \mathbf{e}_{\text{font}} + \mathbf{e}_{\text{char}} \in \mathbb{R}^{d_e}
$$
enabling zero-shot generalization to unseen combinations through linear superposition.
The combined embedding $\mathbf{e}_{\text{total}}$ control the generation process via a scale-shift mechanism used in DiT\cite{DiT}. It is transformed into modulation parameters $\{\alpha, \gamma, \beta\}$ through a MLP with SiLU activation.
This control mechanism significantly reduces computational cost compared to introducing a new text encoder. Furthermore, subsequent experiments demonstrated that this method of control is highly effective.

\section{Experiments}
\label{sec:exp}

\subsection{Dataset}
\begin{figure*}[t]
    \centering
    \includegraphics[width=1\linewidth]{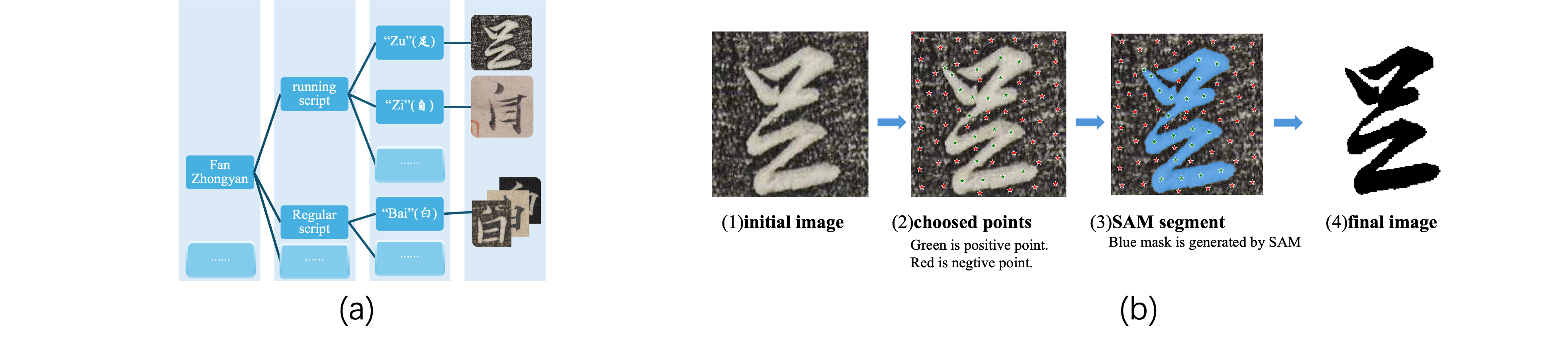}
    \caption{(a) shows the  directory structure of dataset ``Mobao'', using the calligrapher ``Fan Zhongyan'' as an example. ``Zu'' and ``Zi'' only have single images, while ``Bai'' has multiple images. (b) demonstrates the binarization process, using the character ``Zu'' as an example to show the steps of selecting points, obtaining the mask, and resizing.}
    \label{fig:dataset}
\end{figure*}
\textbf{Dataset construct} 
To construct the large-scale annotated dataset ``Mobao'', we scraped images from professional websites, and subject them to binarization processing. 
A total of 1.2 million raw images were initially collected, covering six major categories of Chinese calligraphic scripts: regular script, running script, cursive script, clerical script, seal script, and seal carving. 
Each image in the dataset is meticulously annotated with three critical attributes: calligrapher, font, and character.
The calligrapher labels include 120 prominent figures from Chinese history (e.g., Wang Xizhi, Yan Zhenqing) as well as contemporary practitioners, while font labels strictly follow the six predefined script types.
The dataset is organized into a hierarchical folder structure that mirrors the intrinsic relationships between calligraphers, fonts, and characters. 
Specifically, the top-level directories represent calligraphers, with subdirectories under each corresponding to their associated fonts (e.g., ``WangXizhi\/regularscript\/'' and ``WangXizhi\/cursive\/''). 
Within these font-specific folders, character labels are encoded as subdirectory names (e.g., ``WangXizhi\/regular\/yi\/'' for the character ``yi''), and all images within a character folder are sequentially numbered.
This hierarchical organization not only simplifies data navigation but also inherently encodes the multi-label dependency structure required for conditional generation tasks.
The complete dataset hierarchical is illustrated in Figure \ref{fig:dataset}(a). 

To perform binarization, we employ SAM\cite{SAM} for image segmentation. The complete processing workflow is illustrated in Figure  \ref{fig:dataset}(b). 
This approach addresses the challenges of complex backgrounds and overlapping strokes in historical calligraphy images, which are poorly handled by traditional thresholding methods.
We first apply traditional computer vision techniques to identify connected regions in the image and generate point prompts for SAM.
Initially, we applied Gaussian-filtered Otsu binarization method to roughly separate the foreground from the background. Subsequently, we obtained numerous connected regions from the foreground. 
These connected regions generally fall into two categories: one corresponds to strokes of the calligraphy, and the other consists of noise or dirt spots. 
To distinguish between strokes and noise, we selected a size threshold for the connected regions. 
This threshold was set to 100 pixels: regions larger than 100 pixels were considered as strokes and retained, while those smaller than 100 pixels were classified as noise and removed.
Within these connected regions, positive points are uniformly sampled using the k-means clustering algorithm. Meanwhile, negative points are selected from areas outside of these connected domains.
The number of sampled points is determined according to the following formulas:
\begin{equation}
cnt_{pos} = \max(1,\min(20,\frac{area_{regions} \times 100}{area_{total}})
\end{equation}
\begin{equation}
cnt_{neg} = 50
\end{equation}
Secondly, these sampled points are then used as prompts, together with the original calligraphy images, to guide the Segment Anything Model (SAM). 
The model generates a corresponding segmentation mask, which we interpret as the foreground region of the image. 
To further refine the result, we invert the mask and fill the background region with white pixels, thereby obtaining a clean, binarized image representation.
Finally, a resizing and padding process is performed to standardize the image dimensions. Specifically, the image is first resized such that its longer side is scaled to 256 pixels while preserving the original aspect ratio. 
Subsequently, the resized image is centered within a $ 256 \times 256 $ white canvas by padding the remaining space with white pixels. 
This results in a final square image of size $ 256 \times 256 $ , suitable for downstream processing or model input requirements. \\
\textbf{dataset analyse}
After processing the dataset, we performed a comprehensive analysis of the resulting images. 
A total of 1,929,393 images were obtained, representing calligraphic works from six different fonts, created by 2,681 distinct calligraphers, and covering 4,660 unique Chinese characters.
Notably, some images lacked attribution to a specific calligrapher and were therefore classified under the “anonymous” category. 

To further evaluate the characteristics of our dataset, we conducted a statistical analysis across three key dimensions: calligrapher, font, and character. The results confirm the large scale of our dataset. 
For instance, the calligrapher with the most extensive collection, “Wang Xizhi”, contributes a total of 124,854 images; the font with the largest representation, running script, accounts for 668,040 images; 
and the most frequently occurring character, “shu” (meaning ``book'' in Chinese), appears in 11,949 images.
These statistics demonstrate that the dataset provides a rich and diverse source of data for models to learn both the structural and stylistic features of Chinese calligraphy. 

\begin{figure}
    \centering
    \includegraphics[width=0.98\linewidth]{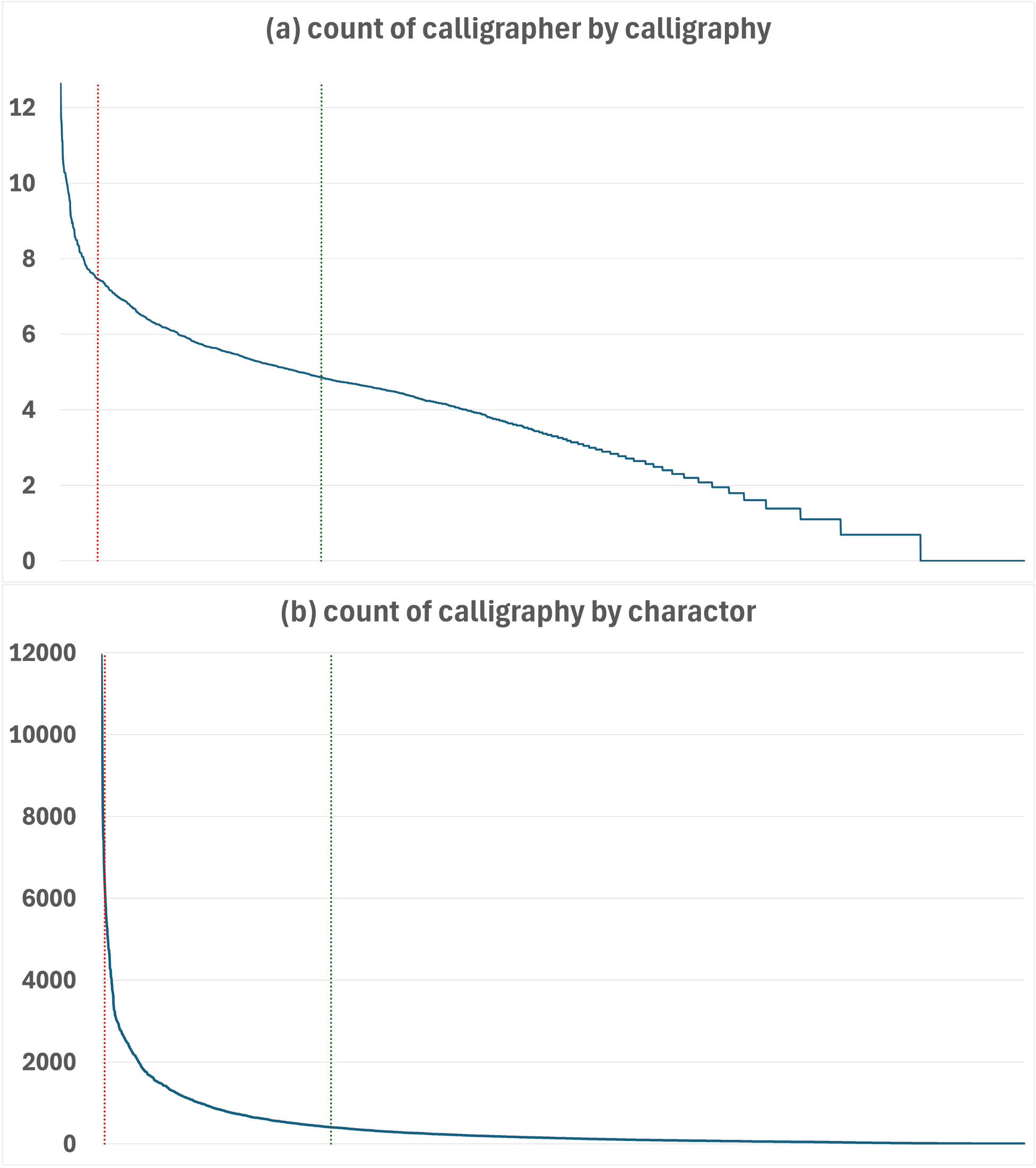}
    \caption{(a) indicates the number of calligraphy per calligrapher.(y-axis:  $\ln$ of calligraphy count) (b) indicates the number of calligraphy per character. Both the red and green lines mark the top $10\%$ and $50\%$ thresholds, respectively, highlighting the long-tail data imbalance.}
    \label{fig:data_distribute}
\end{figure}

However, we also observe notable imbalances in the data distribution.
Nearly half of the calligraphers are associated with fewer than ten images, and approximately half of the characters are represented by less than one hundred samples. 
This suggests that the dataset follows a long-tail distribution as shown in Figure \ref{fig:data_distribute}, where a small number of categories are heavily represented while the majority have limited coverage.
The red and green lines in the figure denote the top $10\%$ and $50\%$ thresholds, further highlighting the skewed nature of the distribution across both calligrapher and character dimensions.
\subsection{Experiment Setup}
\textbf{Experiment Dataset} 
To ensure the reliability and fairness of our experiments, we constructed a balanced subset from the full dataset to serve as the experimental benchmark.
The construction process is as follows: We selected 40 calligraphers who have written a common set of 40 characters.
For each calligrapher, 90\% of the characters were assigned to the training set, while the remaining $10\%$ characters were reserved for testing. 
This ensures that each test character is unseen during training for that specific calligrapher. 
However, other characters written by the same calligrapher, as well as the same characters written by other calligraphers, are still included in the training set. 
This design enables the model to learn both the structural variations of characters and the stylistic characteristics of individual calligraphers.

Given the varying number of available images per character across calligraphers, we randomly sampled up to four images for each character-calligrapher pair.
In total, we obtained 12,985 images, with 11,689 allocated to the training set and 1,296 to the test set.
This careful curation effectively mitigates the negative impact of the long-tail distribution, resulting in a more balanced and representative experimental dataset.
\textbf{Model Specifics}
For the choice of the variational autoencoder (VAE)\cite{VAE}, we employed the same pre-trained VAE as used in Latent Diffusion Models (LDM)\cite{LDM}. 
Specifically, the input image of size $256 \times 256 \times 3$ were first encoded into a lower-dimensional latent space of size $32 \times 32 \times 4$. \\ 
In terms of model configuration, we set the number of iterations \(N\) for our model to 4, with a hidden layer dimensionality of 512. The image segmentation patch size was set to 8. 
During training, we used a fixed learning rate of \(1e-4\) and trained the model on three A100 GPUs with a global batch size of 768. \\
After extensive training, we selected the model checkpoint at 288,000 training steps (approximately 19,199 epochs) for all subsequent experiments.\\

\subsection{Evaluation Metrics}
We evaluated the performance of our model from two complementary perspectives: structural fidelity and stylistic authenticity , which are both critical for high-quality calligraphy generation. \\
To assess the structural accuracy of the generated calligraphy, we employed Tencent's handwritten Optical Character Recognition (OCR) service\footnote{\url{https://cloud.tencent.com/document/product/866/36212}} to recognize the content of each generated image. 
The OCR recognition results were then compared with the ground-truth character labels. 
A higher recognition accuracy indicates that the generated character retains a structure that is close to the original and remains legible to real-world systems. \\
In addition to this semantic-level evaluation, we also adopted two widely used objective metrics—Intersection over Union (IoU) and Peak Signal-to-Noise Ratio (PSNR) —
to quantitatively measure the pixel-level similarity between the generated images and their corresponding ground truth.
These metrics provide an indication of how well the model reconstructs the visual content, reflecting both structural consistency and stylistic resemblance.
Higher values on these metrics generally suggest better alignment in terms of layout, stroke formation, and overall appearance. \\
Beyond automated evaluations, we also conducted out a subjective human assessment to further evaluate the stylistic authenticity and visual quality of the generated calligraphy. 
Specifically, we invited a group of professional calligraphers and art students to evaluate the outputs in a blind test setting.
This qualitative analysis complements the quantitative results by capturing aspects that are difficult to measure with objective metrics, such as artistic expression and resemblance to traditional calligraphic styles.

\subsection{Experiment Results}
\textbf{Qualitative Evaluation}
We employed Tencent's handwritten OCR service to evaluate the structural integrity of the generated calligraphy. 
Specifically, each generated image was submitted to the OCR API, which returned the recognized character. 
This result was then compared with the original character label used during generation. 
If the two matched, the sample was considered to have passed the structural test, indicating that the generated character retained a legible and recognizable form. 

The OCR recognition results are summarized in Table~\ref{tab:OCR}, which reports the number of successfully recognized characters across different font styles. 
For comparison, the table also includes the results obtained on the ground truth (GT) images, as well as the total number of test samples per font category. 
Note that the models and datasets from previous works have not been made publicly available; therefore, we were unable to conduct a direct comparison with those methods. 

\begin{table}[]
\begin{tabular}{l|rrrrr}
\hline
Font        & \multicolumn{1}{l}{regular} & \multicolumn{1}{l}{running} & \multicolumn{1}{l}{cursive} & \multicolumn{1}{l}{clerical} & \multicolumn{1}{l}{seal} \\ \hline
Moyun(Ours) & 219                         & 177                         & 25                          & 17                           & 1                        \\
GT          & 267                         & 240                         & 51                          & 59                           & 5                        \\
Total       & 351                         & 421                         & 347                         & 109                          & 67                       \\ \hline
\end{tabular}
\caption{OCR test result}
\label{tab:OCR}
\end{table}
The results indicate that our model performs well on commonly used fonts like regular script and running script. 
However, the recognition rates for less commonly used fonts such as cursive script, clerical script and seal script were lower. 
This is probably due to the significant differences between these fonts and modern simplified Chinese characters, which caused inaccuracies in OCR recognition. 
We will continue to evaluate structure of the generated calligraphy using other metrics.

We generated the same number of images as the entire test set using the prompts from the test set and evaluated the IOU (Intersection over Union). 
The results showed that our model significantly outperforms other models in terms of IOU. Using the same test set, we also evaluated the PSNR (Peak Signal-to-Noise Ratio) of the model. 
The results demonstrated that our model has a better performance in PSNR either. These findings are presented in Table \ref{tab:2}. 
This indicates that our model is better at fitting the structural integrity of calligraphy characters and more accurately replicating the style of calligraphy. 
\begin{table}
    \centering
    \begin{tabular}{c|cc}
    \toprule
        Method & IOU$\downarrow$ & PSNR$\uparrow$\\
         \hline
         CalliGAN & 0.325 & -  \\
        ZiGAN & 0.348 & -  \\
        Zipeng Zhao\cite{zhou_end--end_2021} & - & 25.3734  \\
        Moyun(Ours) & 0.260 & 32.0727  \\
    \bottomrule
    \end{tabular}
    \vspace{0.1cm} 
    \caption{Qualitative measurements}
    \label{tab:2}
\end{table}
\\
\textbf{Quantitative Evaluation}
To rigorously assess the performance and stylistic fidelity of our generated calligraphy, we designed a human evaluation questionnaire. This questionnaire aimed to gauge how closely our model's outputs resemble authentic works from specific calligraphers, thereby providing an objective measure of the model's ability to capture and reproduce nuanced artistic styles.

\begin{figure}
    \centering
    \includegraphics[width=1\linewidth]{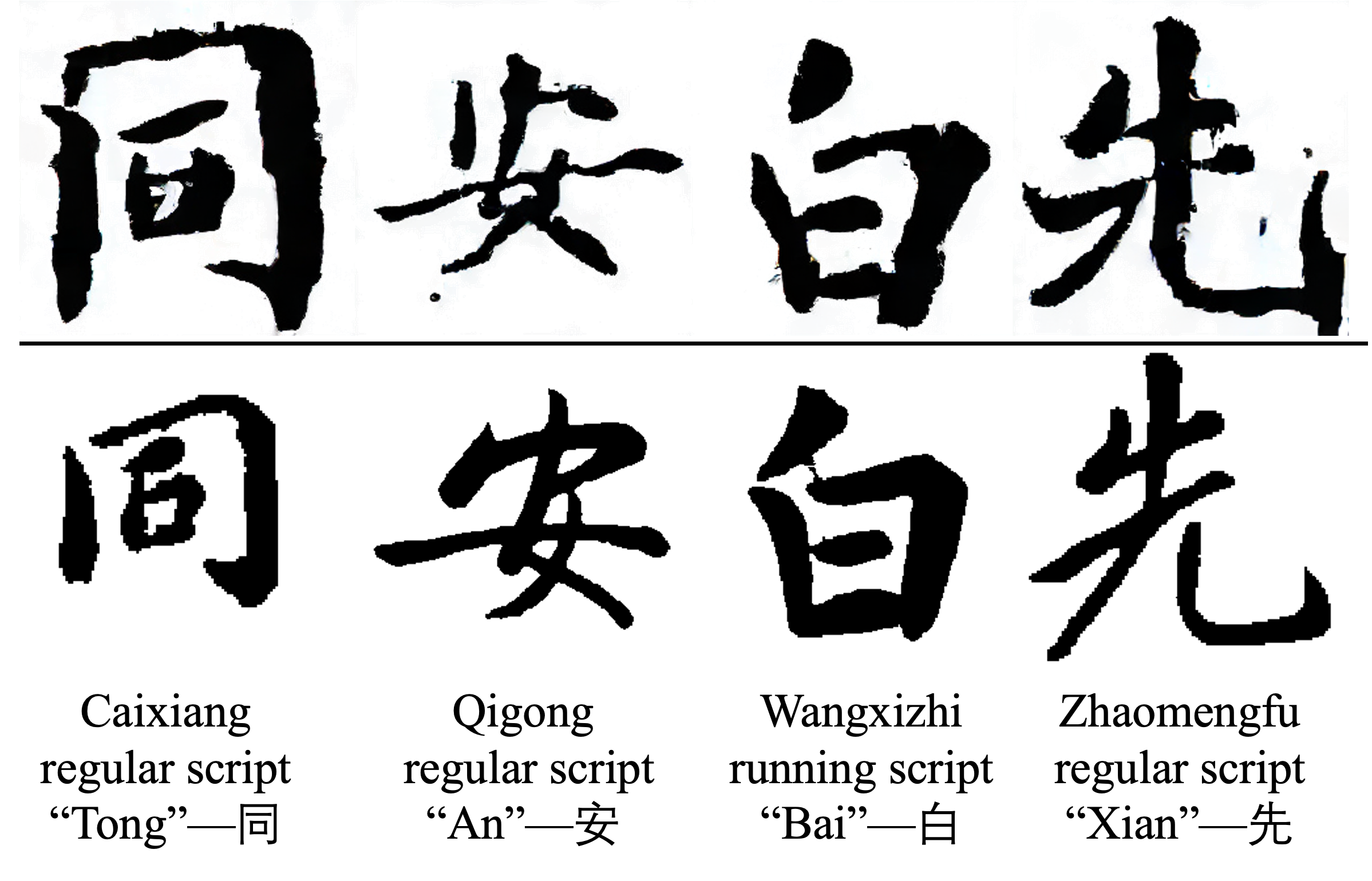}
    \caption{Generated calligraphy. Each column with different labels. The first row shows calligraphy generated by the model which were unseen before. and the second row is ground truth.}
    \label{fig:ans}
\end{figure}

We selected ten representative calligraphy samples generated by our model, each corresponding to a distinct character. These samples were chosen based on their visual quality and stylistic diversity, ensuring a comprehensive evaluation across various aspects of calligraphic artistry.
Some of these calligraphy was shown in Figure \ref{fig:ans}. 

In our questionnaire, each question is based on a calligraphy image generated by our model, which serves as the prompt. Below the prompt, four options are presented, each showing a real-world calligraphy sample of the same character written by a different renowned calligrapher. Among these four options, only one corresponds to the same TripleLabel used to generate the prompt image—this option is referred to as the "Backbone" option. Participants are asked to select the sample that most closely matches the style of the generated image. This design allows us to evaluate whether the generated results are perceived as stylistically consistent with the intended real calligrapher.
The questionnaire was administered to 9 participants, including both calligraphy enthusiasts and individuals with general interest in Chinese art, to evaluate the stylistic fidelity of our generated calligraphy. Each participant was presented with 10 questions, resulting in a total of 90 responses. The results show that $53.3$\% of the generated calligraphy samples were correctly paired with their corresponding ground-truth style—i.e., participants selected the "Backbone" option (the real calligrapher matching the TripleLabel) as the most stylistically similar sample.

This accuracy rate, significantly above the random guessing baseline of $25$\% (chance level for four choices), indicates that our model successfully captures and reproduces key stylistic features of the target calligraphers. It suggests that the generated characters are not only visually plausible but also semantically aligned with the intended artistic style, to the extent that human observers can reliably detect the stylistic correspondence despite the subtle differences inherent in calligraphic expression.

Furthermore, the result supports the effectiveness of our approach in leveraging the TripleLabel to condition the generation process, enabling fine-grained control over stylistic output. While there remains room for improvement—particularly in handling highly nuanced or idiosyncratic styles—this human evaluation provides strong qualitative evidence that our model generates perceptually coherent and stylistically consistent calligraphy, bridging the gap between synthetic generation and authentic artistic expression.
The models and datasets of other works have not been open-sourced, so they were not included in the questionnaire.
\section{Conclusions}
\label{sec:conclusion}

In this paper, we proposed a new calligraphy generation model, ``Moyun'', which could generate calligraphy in a specified style guided by the three labels: calligrapher, font, and character. The core idea was the introduction of Vision Mamba and the development of the TripleLabel control method. Additionally, we collected a large-scale, well-annotated, and properly binarized calligraphy dataset ``Mobao'', which further demonstrated the effectiveness of our work.

\begin{acks}
This work was supported by  the Shanghai Municipal Science and Technology Commission (24511106802), and the computation is performed in ECNU Multifunctional Platform for Innovation(001).
\end{acks}

\bibliographystyle{ACM-Reference-Format}
\balance
\bibliography{moyun-ref}










\end{document}